\documentclass[letterpaper, 10 pt, conference]{ieeeconf}  

\IEEEoverridecommandlockouts                              

\usepackage{graphicx}
\usepackage{caption}
\usepackage{subcaption}
\usepackage{lipsum}
\usepackage{booktabs}
\usepackage{multirow}
\usepackage{hyperref}
\pdfoutput=1

\title{\LARGE{\bf{HoloSpot: }}\Large{\bf{Intuitive Object Manipulation via Mixed Reality Drag-and-Drop}}
}

\author{Pablo Soler$^{1,\star}$, Petar Lukovic$^{1,\star}$, Lucie Reynaud$^{1}$, Andrea Sgobbi$^{1}$, Federica Bruni$^{1}$, Martin Brun$^{1}$, Marc Zünd$^{1}$, \\ Riccardo Bollati$^{1}$, Marc Pollefeys$^{1, 2}$, Hermann Blum$^{1, 3, \dagger}$ and Zuria Bauer$^{1, \dagger}$ \\
{\textit{$^{1}$ETH Zurich $^{2}$Microsoft $^{3}$Lamarr Institute / Uni Bonn}}
\thanks{$^{\star}$Shared first authorship. $^{\dagger}$Equal supervision. This project was partially funded by the ETH Foundation. We thank Jiaqi Chen for providing guidance to develop the HoloLens app and all our participants for the user study.}\\
\tt\small \{psoler, plukovic, lreynaud, asgobbi, fbruni, mabrun,  mazuend, rbollati \\ 
\tt\small pomarc, blumh, zbauer\} @ethz.ch}%

\begin{document}
\maketitle

\begin{abstract}
Human-robot interaction through mixed reality (MR) technologies enables novel, intuitive interfaces to control robots in remote operations. Such interfaces facilitate operations in hazardous environments, where human presence is risky, yet human oversight remains crucial. Potential environments include disaster response scenarios and areas with high radiation or toxic chemicals.
In this paper we present an interface system projecting a 3D representation of a scanned room as a scaled-down 'dollhouse' hologram, allowing users to select and manipulate objects using a straightforward drag-and-drop interface. We then translate these drag-and-drop user commands into real-time robot actions based on the recent Spot-Compose framework. The Unity-based application provides an interactive tutorial and a user-friendly experience, ensuring ease of use.
Through comprehensive end-to-end testing, we validate the system's capability in executing pick-and-place tasks and a complementary user study affirms the interface's intuitive controls. Our findings highlight the advantages of this interface in improving user experience and operational efficiency. This work lays the groundwork for a robust framework that advances the potential for seamless human-robot collaboration in diverse applications. Paper website: \href{https://holospot.github.io/}{\texttt{https://holospot.github.io/}}.
\end{abstract}



\section{INTRODUCTION}
Human-robot interaction (HRI) is becoming more and more important due to increasing deployment of robots and the complementary nature of human and machine capabilities. Robots are now commonly deployed in scenarios where human presence is risky or impractical, such as hazardous environments and remote locations. Examples include industrial settings with toxic materials, areas affected by natural disasters, and space exploration missions. These deployments minimize the need for on-site human intervention, enhancing operational efficiency and safety~\cite{Szczurek2023, CERN2023}.

A significant challenge in HRI lies in the interfaces used to control robots. Traditionally, three-dimensional (3D) environments are represented on two-dimensional (2D) screens. This representation can complicate the operator's ability to maintain situational awareness and precision by limiting its depth perception and field of view. This makes it especially challenging to interact with and manipulate objects accurately in a 3D space~\cite{Ahlberg2022}. Mixed reality (MR) technology offers a promising solution by providing a more immersive and intuitive interface for human-robot interaction. This approach addresses the limitations of traditional 2D interfaces by providing a 3D visual representation that aligns with the real-world environment for better scene understanding.



\begin{figure}[t]
\centering
\vspace{-2mm}
\includegraphics[width=1.0\linewidth]{ 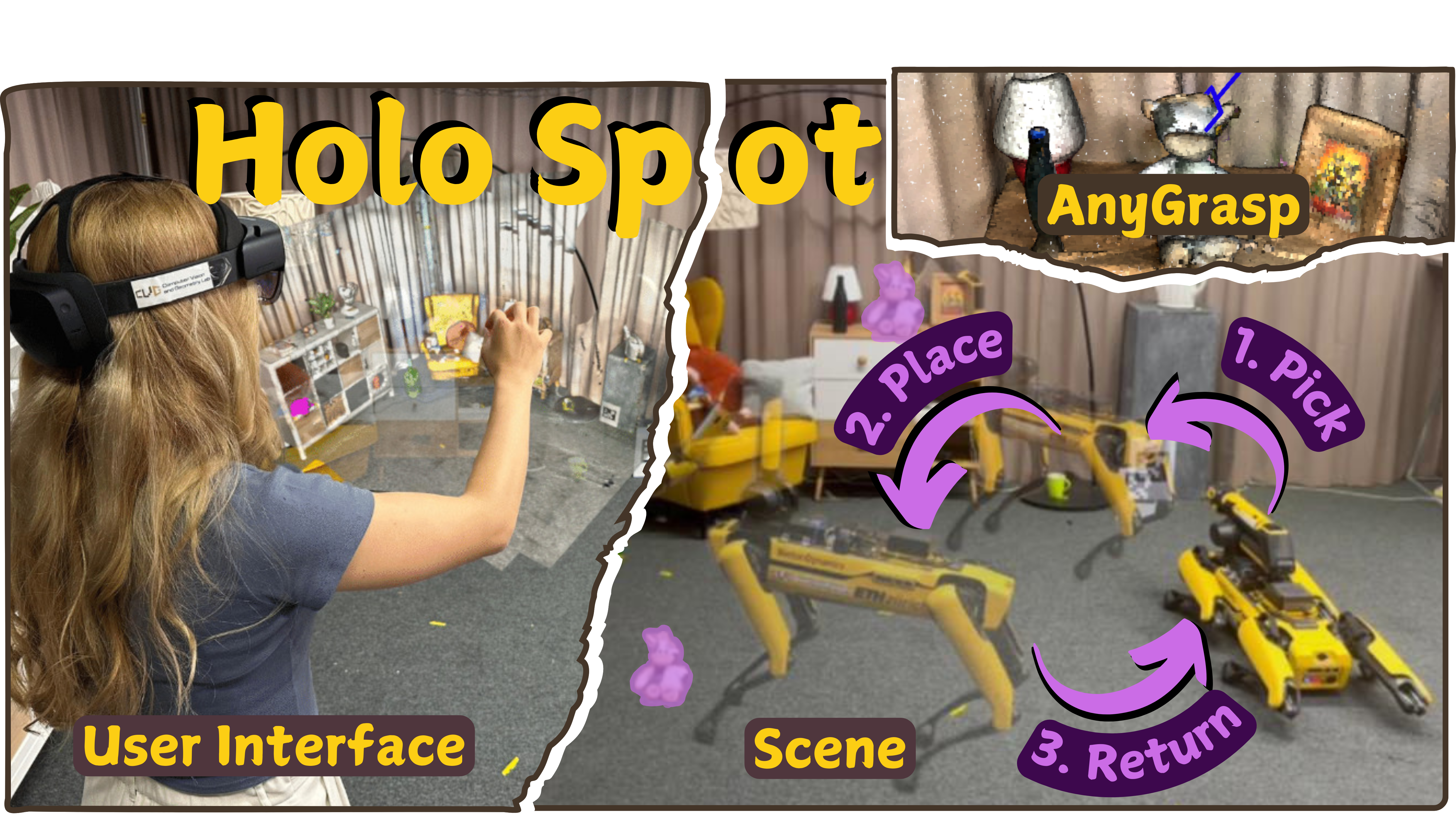}
\vspace{-6mm}
\captionof{figure}{\textbf{Visualization of Our System.}
Left: A user wearing a Hololens to interact with a holographic representation of the environment. Right: Robot pick-and-place sequence of actions. Top: Optimal grasp selection with AnyGrasp.}
\label{fig:starting_image}
\vspace{-0.5cm}
\end{figure}

Recent research has explored the use of MR headsets to control robots. For example, Chen et al. demonstrate real-time visualization and intuitive navigation of robots in 3D, highlighting the benefits of MR interfaces over traditional methods~\cite{chen20233d}. These studies underscore the advantages of MR technology in enhancing the interaction between humans and robots, making it possible to perform more complex and precise tasks.



Building on this, our project presents an innovative design for a user-friendly MR application, enabling users to command a robot to perform pick-and-place tasks. The primary goal of our project is to create an interface that allows users to control a mobile object manipulator through natural, intuitive actions in a MR environment. 


Our system achieves this by implementing several key features. First, we scan and reconstruct the deployment environment in 3D. Second, we segment the reconstruction into individual object instances through 3D object instance segmentation~\cite{schult2023mask3d},~\cite{takmaz2023openmask3d}. Third, we make the digital twin interactive based on the object segmentation and link any interaction in the digital twin to a robot that reproduces the same environment interaction in the real environment. Therefore, by simply dragging and dropping virtual objects, users can instruct the robot to pick up and move these objects to specified locations in the real world.
To ensure ease of use, our application includes an interactive tutorial that guides users through the interface and functionality. This tutorial helps users quickly get familiar with the system, making it accessible even to those with minimal technical expertise. 

The main contributions of this work are:
\begin{itemize}
    \item We present a novel interface for remote robot operations that enables object manipulation using a drag-and-drop feature within a MR environment.
    \item We implement the proposed interface as an app for the HoloLens and the Spot, featuring an interactive tutorial, movable objects, and operable drawers.
    \item We perform extensive real-world tests with our interface and a user study, demonstrating that our interface is intuitive and enables users to perform complex pick-and-place tasks with minimal training.
\end{itemize}
\vspace{-0.1cm}

\section{RELATED WORKS}
The intersection of Mixed Reality and Human-Robot Interaction has become a dynamic research area, with studies exploring the potential of MR to enhance robotic control and human collaboration in complex environments. We highlight a selection of recent and particularly pertinent studies.

\begin{figure*}[t]
\vspace{-0.5cm}
\centering
\includegraphics[width=\linewidth]{ 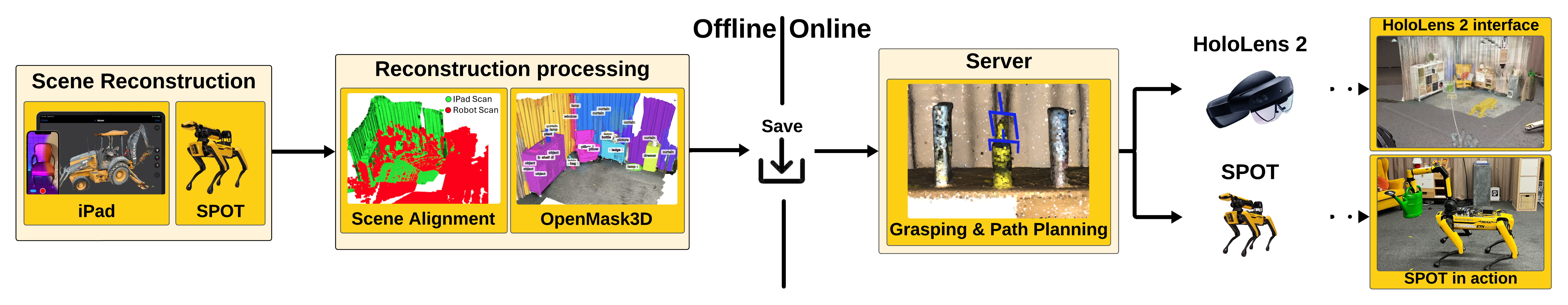}
\vspace{-7mm}
\caption{\textbf{System overview.} 
Our method relies on both offline and online segments. Offline segment (left), is used to construct 3D scene that will be utilised in the subsequent online phase. The online segment (right) is used to control the Spot using the HoloLens with the help of an online server.}
\label{fig:system_overview}
\vspace{-0.5cm}
\end{figure*}



\subsection{Control via Mixed Reality Interfaces}

Human-Robot Interaction (HRI) has been transformed by recent advances in Mixed Reality (MR) technologies.
specially in intuitive control and situational awareness. Early work, such as~\cite{chen20233d}, showcased how 3D visualization in MR enhances real-time robotic system control, bridging the gap between abstract mechanisms and direct manipulation. This intuitive spatial understanding set the foundation for later developments.
%
Alternatively,~\cite{iglesius2024mrnabmixedrealitybasedrobot} delved into robotic navigation with MRNaB, leveraging MR beacons to improve human-robot cooperation in intricate spatial settings.

In parallel,~\cite{Suzuki_2022} categorized AR's influence on HRI, with particular attention to its advantages in enhancing teleoperation and task management in complex environments. This contributes to ongoing MR interface research that seeks to enhance precision and fluidity in interaction. \cite{Digital_twin_driven_MR} pushed these developments further by incorporating digital twins and haptic feedback, adding tactile input to the visual interface. This system improves control precision, especially in scenarios where visual data alone is insufficient. Their work represents a significant evolution in MR, highlighting its potential to create more effective HRI systems.


\subsection{3D Instance Segmentation and Grasp Pose Estimation}

One of the main challenges in robotics lies in understanding and interacting with complex environments. This issue is often addressed via 3D instance segmentation and grasp pose estimation.
The Spot-Compose framework~\cite{lemke2024spotcompose} tackles this by using point cloud segmentation algorithms to identify objects, making it highly effective for tasks like object retrieval and manipulation in dynamic environments. In parallel, OpenMask3D~\cite{takmaz2023openmask3d} enhances 3D segmentation by employing open-vocabulary models. This enables robots to interact with previously unknown objects based on natural language descriptions. This flexibility is essential for improving adaptability in real-world applications.

Simultaneously, grasp pose estimation has seen substantial progress through the development of AnyGrasp~\cite{fang2023anygrasp}. AnyGrasp predicts two-finger grasps directly from 3D point clouds, using a dense supervision strategy that incorporates real perception data and analytic labels. This method accounts for factors such as the object's center-of-mass and environmental constraints, ensuring stable and collision-free grasps. Grasp pose estimation has progressed significantly, with frameworks like GraspNet-1Billion~\cite{fang2020graspnet} which predicts stable, collision-free grasps from 3D point clouds by analyzing object geometry and environmental constraints. This approach complements the capabilities of Spot-Compose and OpenMask3D in segmentation tasks. Moreover,~\cite{duan2021grasping} explored techniques based on deep learning to enhance grasping in robotic manipulation, improving the handling of objects under various physical conditions.

\subsection{Enhancing Human-Robot Collaboration through MR}

Human-robot collaboration is critical in environments requiring precision, situational awareness, and real-time decision-making. MR interfaces have revolutionized this collaboration by enabling seamless interaction between human operators and robotic systems. Research by~\cite{erat2018drone} underscores MR’s versatility in improving remote control of robotic systems, particularly in environments where human presence is limited but precision is important. This approach enables operators to manage tasks more effectively, enhancing both performance and safety.

Augmented reality (AR) interfaces, as investigated by~\cite{mott2021augmented}, have shown promise in improving situation awareness in human-robot teaming for exploring tasks. Their study evaluates how AR visualizations can support better decision-making during time-sensitive operations, questioning about the effectiveness of these tools in collaborative settings. Furthermore,~\cite{bejczy2020mixed} explored the utility of MR interfaces in contamination-critical environments. It shows how such systems enhance teleoperation accuracy and reduce error rates.

Similarly,~\cite{delmerico2022spatial} explored how MR can optimize human-robot interaction (HRI) in remote manipulation tasks, significantly lowering cognitive load while improving task precision. Collaborative MR workspaces, like those developed by~\cite{Szczurek2023}, allow multiple users to control and monitor robotic systems simultaneously. This shared interface improves coordination and task efficiency, particularly in team-based operations requiring collaborative control.

Finally,~\cite{reardon2020situational} developed a system where autonomous robots detect changes in the environment and communicate these through augmented reality, enhancing human situational awareness. Their work highlights the potential of AR in supporting human-robot teams by making environmental data more accessible and understandable.

\section{METHOD}

In this section, we present the main components and techniques employed by the system. An overview is illustrated in Figure~\ref{fig:system_overview}. In the following we go through each part of our proposed method.

\vspace{-0.2cm}

\subsection{Obtaining 3D Scene Reconstruction}


We assume an available pre-scanned 3D representation of the scene. This representation can be obtained through SLAM with a suitably equipped robot or handheld scanners. In our experiments, we use a LiDAR-equipped iPad and the \textit{3D Scanner App} ~\cite{3dscan} (high detail scan). Additionally, like in~\cite{lemke2024spotcompose}, we instruct the robot to perform a depth scan (low detail scan) with its onboard cameras. While the robot's cameras face downward and therefore cannot be used alone to reconstruct the scene, the two registered point clouds then form a comprehensive 3D reconstruction of the scene, which is used in all subsequent steps. 


\vspace{-0.2cm}

\begin{figure}[t]
    \centering
    \includegraphics[width=1.0\linewidth]{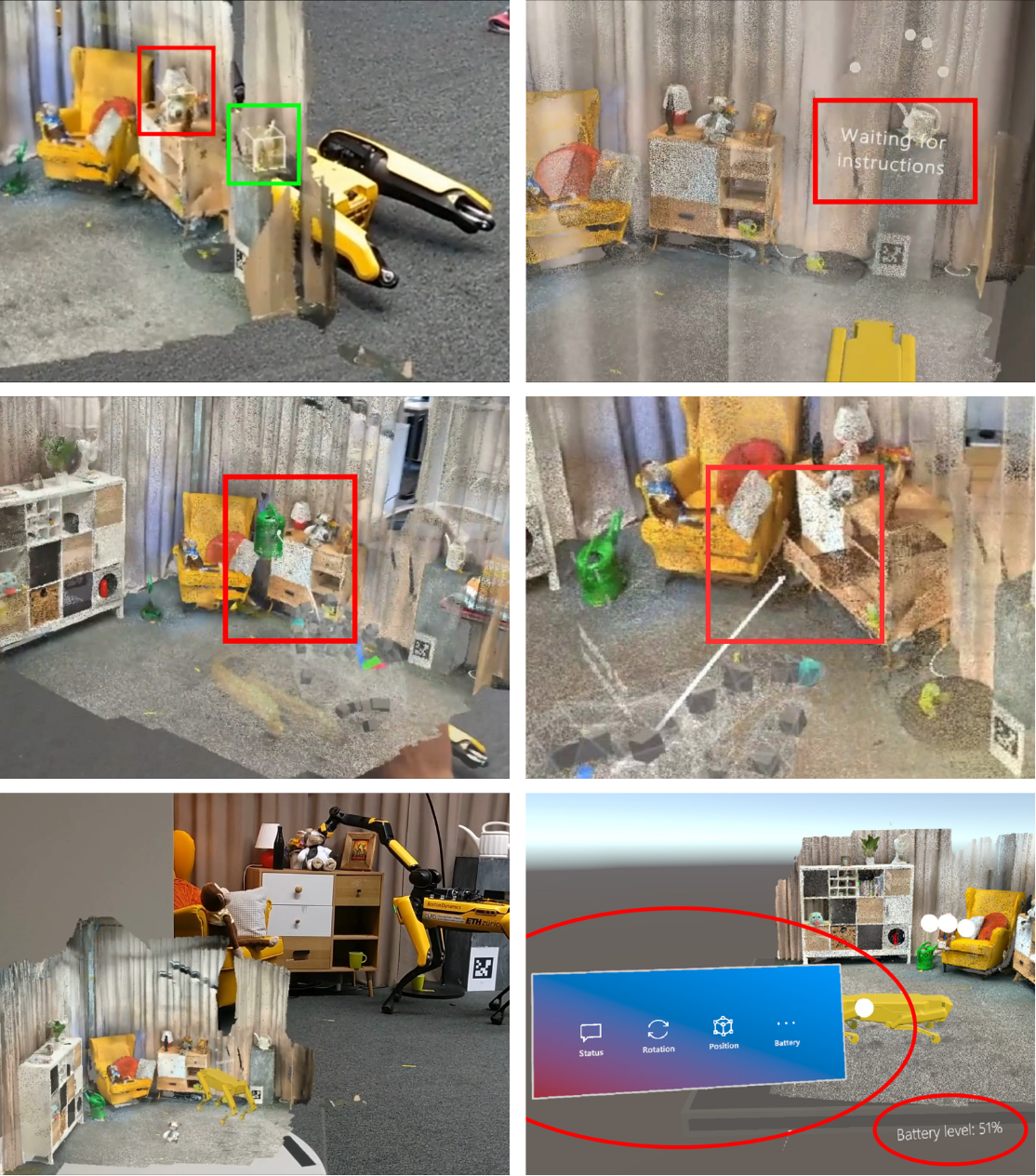}
    \caption{\textbf{HoloLens Interface.} The above figures show various aspects of the visual interface on HoloLens. Labeling images with number left to right, and top to bottom we have: on the first image bounding boxes displaying movable objects following "show items" voice command, the following image shows how robot status is displayed during manipulation. Images four and five display user manipulating watering can and a drawer. On the fifth and sixth images we can see virtual representation of the robot and menu containing battery percentage and additional status information.}
    \label{fig:hololens_interface}
    \vspace{-0.6cm}
\end{figure}

\subsection{3D Semantic Instance Segmentation}
\begin{figure*}[t]
\vspace{-0.5cm}
\centering
\includegraphics[width=\linewidth]{ 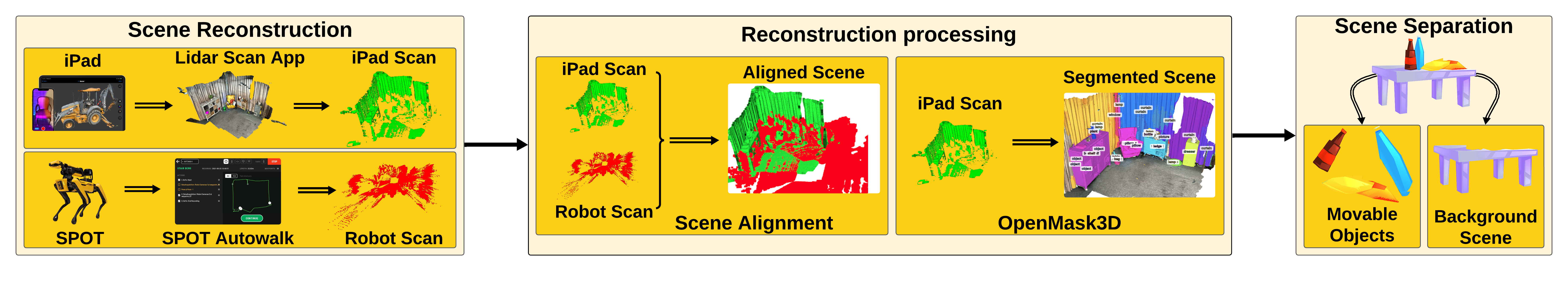}%
\vspace{-5mm}
\caption{\textbf{Offline pipeline.} 
The preprocessing done before deploying our system can be separated into three parts: Scene Reconstruction, Reconstruction Processing and Scene Separation. Scene reconstruction consists of gathering high and low resolution scans using the iPad LiDAR and the Spot cameras, respectively. Recorded point clouds are then aligned into the same coordinate system and the high resolution scan is segmented using OpenMask3D~\cite{takmaz2023openmask3d}. At the end we manually separate segmented instances into draggable objects and static environment.} 
\label{fig:offline_pipeline}
\vspace{-0.5cm}
\end{figure*}

We segment the 3D reconstruction into semantic instances using the OpenMask3D~\cite{takmaz2023openmask3d} framework. This is illustrated in subsequent steps in Figures~\ref{fig:system_overview} and~\ref{fig:offline_pipeline}. This state-of-the-art method supports open vocabulary queries on 3D scenes, allowing us to predict and delineate object instances within a 3D point cloud together with their semantic descriptors. 

The result of this step is presented at the end of the Figure~\ref{fig:offline_pipeline}. The point cloud is segmented into each movable object instance and the static environment. It is worth noting that we provide a predefined list of language prompts to select movable objects, or segmented point cloud for that matter. All other parts of the scene are considered static.


\subsection{Unity App Interface on the HoloLens}

To enable intuitive interaction with the reconstructed 3D environment, we develop a Unity application (see Figure~\ref{fig:hololens_interface}) for the Microsoft HoloLens~2~\cite{hololens}. The interface shows the 3D reconstruction as a hologram in front of the user, where they can then with their hands drag-and-drop the segmented movable objects through the HoloLens hand tracking.

The application imports the  point cloud of the scene, with each movable object instance as an independent grabbable hologram. To facilitate user manipulation, we further add a virtual floor to the hologram. When an object hologram is grabbed and subsequently dropped, custom scripts send the object's new coordinates to the main workstation. To ensure operational feasibility, a control mechanism verifies whether the drop location is within the robot's operational area. 

During testing, we found it beneficial to include a visual representation of the robot with a status display to keep users informed of the robot's activities. For users unfamiliar with the HoloLens, we introduce a tutorial to explain the app's features and practice drag-and-drop. We also add confirmation buttons to avoid sending wrong object placements to the robot and to alert users if the chosen location is not feasible. Additionally, we integrate voice control features: the \textit{"show items"} command highlights movable objects to the user with colored bounding boxes, while the \textit{"reset"} command allows users to return all moved objects to their original positions. All features are illustrated in Figure~\ref{fig:hololens_interface}.

\subsection{Online operation of the Robot}

\begin{figure*}[h]
\centering
\includegraphics[width=0.9\linewidth]{ 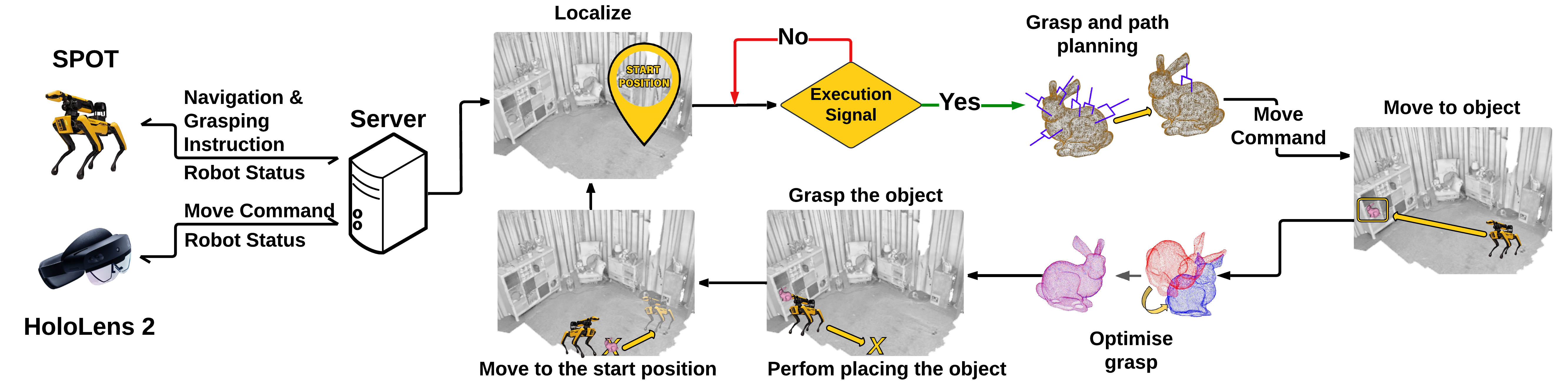}
\vspace{-1mm}
\caption{\textbf{Online pipeline.}
When the system is deployed (online) it follows the given loop. Start of the pick-and-place procedure is triggered by the HoloLens user when he places an object in the scene. This sends a signal to the intermediate server with additional information about the object and its location. After a successful information exchange, the robot is localized. Next, the grasp and path are calculated on the server which then sends the commands to the robot. After the robot arrived to the location, grasp optimisation is performed using ICP algorithm \cite{121791}. At the end the robot performs the grasp, moves the object and returns to the starting position where it localizes itself waiting for another trigger signal.}
\label{fig:online_pipeline}
\vspace{-0.6cm}
\end{figure*}

After an object is selected by the user and dropped at a new location in the interface, the interaction gets translated into a pick-and-place command and sent to the robot. In Figure~\ref{fig:online_pipeline}, this is depicted as \textit{"Navigation \& Grasping Instruction"}. The command includes the drop coordinates in the 3D scene and the index of the object in the scene representation that is available to both the robot and the interface.

To execute the pick-and-place command, we use the Spot-Compose~\cite{lemke2024spotcompose} framework that we briefly outline here: Picking and placing starts with grasp estimation on the object using AnyGrasp~\cite{fang2023anygrasp}. We run inference over multiple rotations of the instance mask, since AnyGrasp identifies poses based on the frontal view. The system then performs joint optimization of poses and grasps based on the AnyGrasp \cite{fang2023anygrasp} score, the alignment of the robot body with the grasp pose and the vicinity of obstacles. Once the robot has moved to the pose from which it should grasp the object, a local point cloud of the object is captured using the depth camera located in the robot gripper. This local capture is aligned to the initial scan using ICP~\cite{121791} to obtain a corrective transformation for the grasp pose, compensating any misalignment or drift between robot odometry and scene representation (see \textit{"Optimise grasp"} in Figure~\ref{fig:online_pipeline}).
Next, the robot grasps the item, picks it up, and subsequently moves to the drop location. The route to this location is calculated at the same time as the route from start to the object location, both routes are calculated using RRT~\cite{LaValle1998RapidlyexploringRT}. We also run joint optimisation for drop location (see \textit{"Optimise grasp"} in Figure~\ref{fig:online_pipeline}). This is computed in parallel during the pick operation, since the drop location is known a priori. 
Finally, Spot drops the object and moves back to the start location and localizes itself again to alleviate drift.

\vspace{-0.1cm}

\subsection{System Integration}

Our system is based on a centralized architecture, with all devices connected via WiFi (these include the Agile Mobile Robot Spot~\cite{spot}, the Microsoft HoloLens~2~\cite{hololens} and the server). The server is responsible for the entire planning procedure and sends commands to the Spot robot using the Boston Dynamics Python SDK~\cite{bosdyn}. The AnyGrasp~\cite{fang2023anygrasp} and OpenMask3D~\cite{takmaz2023openmask3d} models also run on this machine.
The HoloLens app queries the robot status and issues commands via a REST API exposed by the workstation.

\vspace{-0.1cm}

\subsection{Drawer Drag-and-Drop}
To showcase the versatility of the proposed interface, we implement interactable drawers that can be opened and closed using the same drag-and-drop mechanism. Their movement is constrained to match the range of motion of real-world drawers. This is combined with the Spot Compose functionality to allow the robot to open drawers. An example of the user interacting with the drawer is displayed in the fourth image of Figure~\ref{fig:hololens_interface}.

\vspace{-0.1cm}

\section{EXPERIMENTS}
To evaluate our system, we first evaluate the end-to-end reactivity and reliability of our proposed interface. This is to validate that the interface does not introduce critical additional points of failure to the mission deployment of the mobile manipulator. We then conduct a user study to compare the effectiveness of our proposed interface with a much more common screen + mouse interface.

All the experiments were performed using the Microsoft Hololens 2~\cite{hololens} for the MR interface and the Agile Mobile Robot Spot~\cite{spot}. For computing the grasping and path planning we used a workstation with an Nvidia RTX 4090. The commands are then sent to a device that is mounted on the robot and communicates with the Spot via ROS~\cite{SPOTROS}. All communications are done with a router using static IPs.

\subsection{Real World System Test}

\begin{figure}[h]
    \centering
    \includegraphics[width=1.0\linewidth]{ 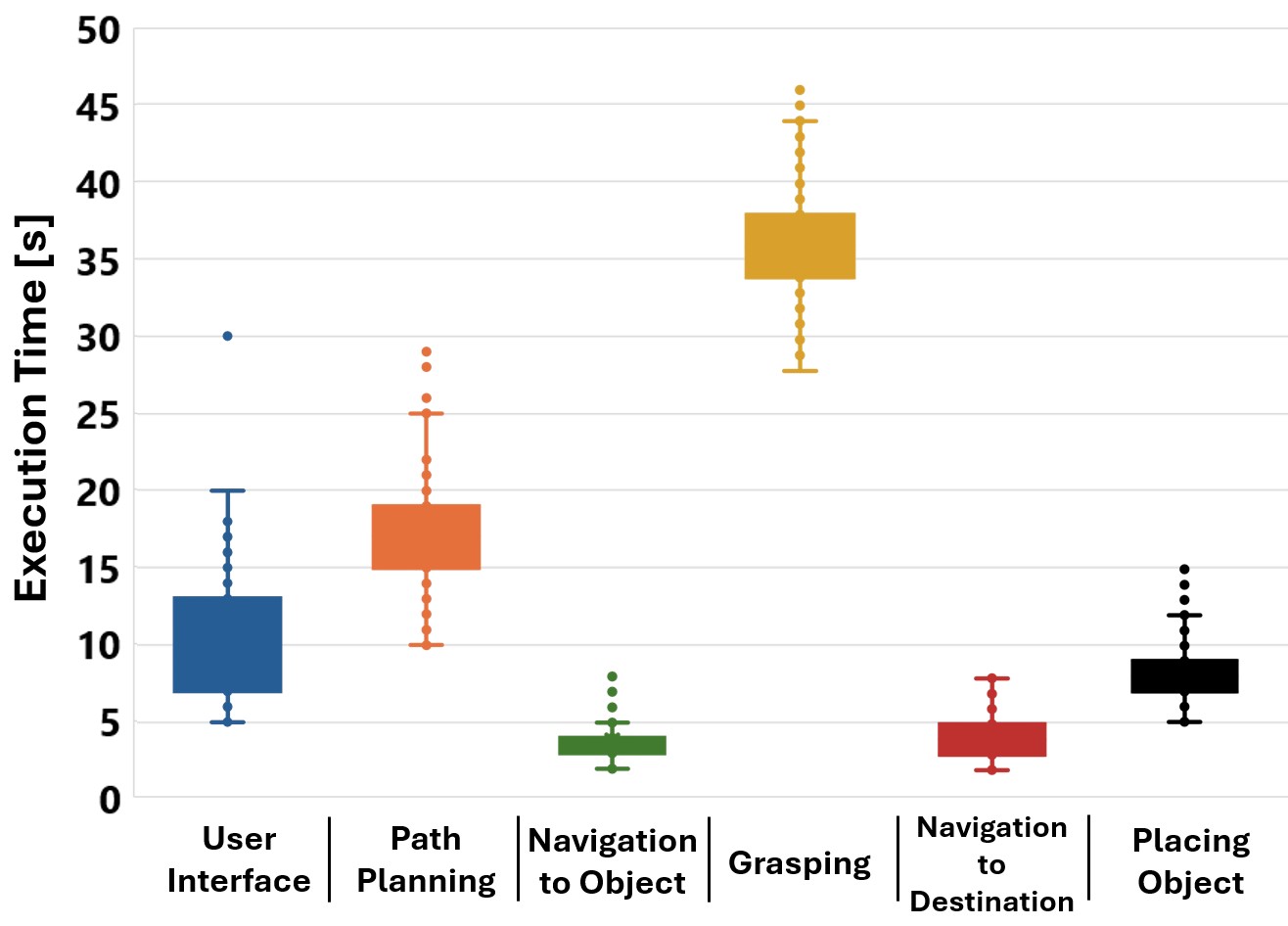}
    \vspace{-5mm}
    \caption{\textbf{Task Completion Time Breakdown.}
    Mean execution times for the six stages of object placing: user interface, path planning, navigation to object, grasping, navigation to destination and placing object, respectively. Data from 240 trials; error bars show ±1SD and ±3SD.}
    \label{fig:execution time}
    \vspace{-3mm}
\end{figure}

We run 240 pick-and-place actions using a variety of objects of different difficulties: Two differently sized watering cans, a mug, two toy plushies and differently shaped and colored bottles. In Table~\ref{tab:manipulation_stats_table}, the different pick up locations are listed. During these trials, the robot is instructed through our HoloLens interface to pick up a specified object. Each object is moved 40 times from its original position to a fixed dropping place. For a more detailed evaluation, we divide each task in phases, which are shown in Figure~\ref{fig:execution time}. The two main metrics used for the real world tests are: Success rate and task duration. Success rate is responsible for measuring reliability of our system, whereas, task duration gives us a way to measure user experience and optimize our system.

\begin{table}[h]
    \vspace*{4pt}
    \centering
    \begin{tabular}{@{}p{2cm}p{1.2cm}p{0.9cm}p{3.5cm}@{}}
    \toprule
    \textbf{Object} & \textbf{Pick-up\newline Location} & \textbf{Success Rate} & \textbf{Main Issue} \\
    \midrule
    White Can & Shelf & 75\% & Collision on navigation\\
    Green Can & Floor & 85\% & Collision on navigation\\
    Green Mug & Floor & 70\% & Empty Grasp \\
    Black Bottle & Cabinet & 80\% & Object not found \\
    Blue Plush & Cabinet & 62\% & Collision during grasp \\
    Cow Plush & Cabinet & 55\% & Collision during grasp \\
    \bottomrule
    \end{tabular}
    \caption{\textbf{Object Manipulation Overview.} The different objects help to identify critical failure modes in the underlying mobile manipulation system.}
    \label{tab:manipulation_stats_table}
    \vspace*{-8pt}
\end{table}

\textbf{Task Duration analysis:} According to the data shown in Figure~\ref{fig:execution time}, the total average time of end-to-end task execution is 80 s. 
17 s are taken for the path and grasp planning, which is the first step after a command is issued and therefore the most critical delay between user input and robot action. The user interaction through the drag-and-drop interface typically takes about 10 seconds, showing the effectiveness and user-friendliness of our application. 

\begin{figure*}[t]
    \vspace*{-30pt}
    \centering
    \includegraphics[width=0.8\linewidth]{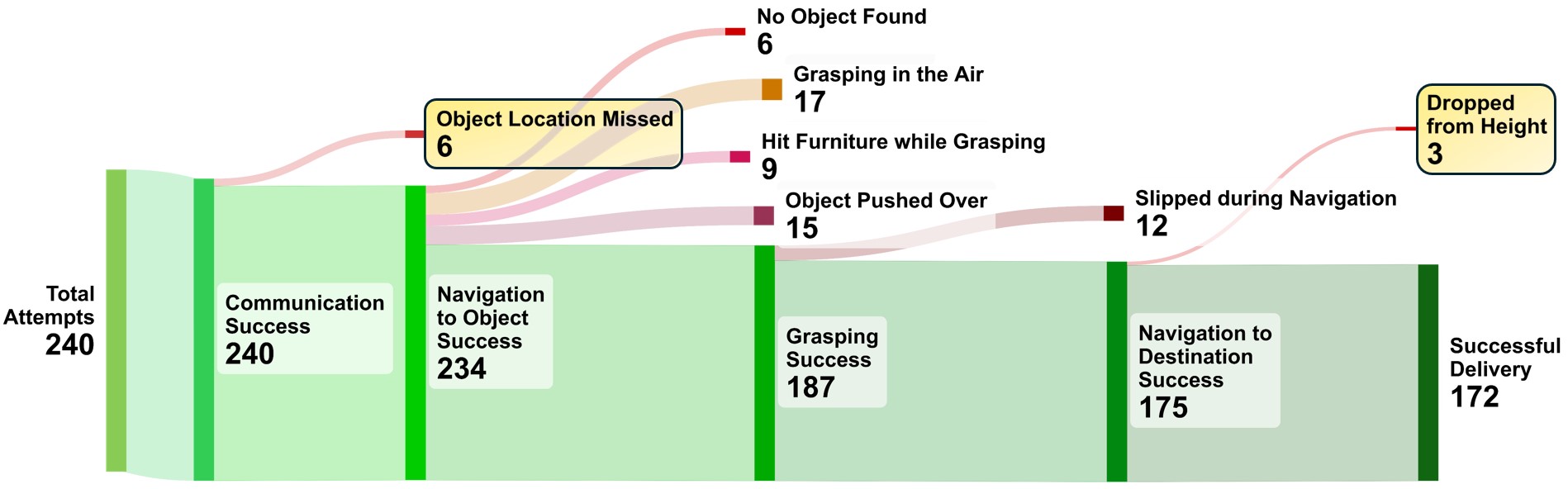}
    \vspace{-2mm}
    \caption{\textbf{Manipulation Experiments Results.}
    To assess the object manipulation reliability of our interface, we conducted 240 trials using six distinct objects, differing in grasping difficulty and pick-up location. The overall success rate is 72\%. The most critical part is the object grasping and the primary failure is missing the object and grasping in the air. Failures highlighted in yellow are directly linked to our interface.}
    \label{fig:manipulation-success}
    \vspace*{-15pt}
\end{figure*}

\textbf{General pick-up success:} As shown in Figure~\ref{fig:manipulation-success}, on 240 trials, 172 (72\%) were successful. This is overall in line with the findings from the mobile manipulation system that we use~\cite{lemke2024spotcompose}. Notably, from 68 failure cases, 9 (13\%) can be attributed to the user interface due to inaccurate pick-up or drop-off commands. The other failures come from the grasping (47) and navigation (12) systems, which are independent of our proposed interface. We conclude that the proposed interface does not drastically increase failure modes of the overall robotic system. 

\begin{figure}[t]
    \centering
    \includegraphics[width=0.9\linewidth]{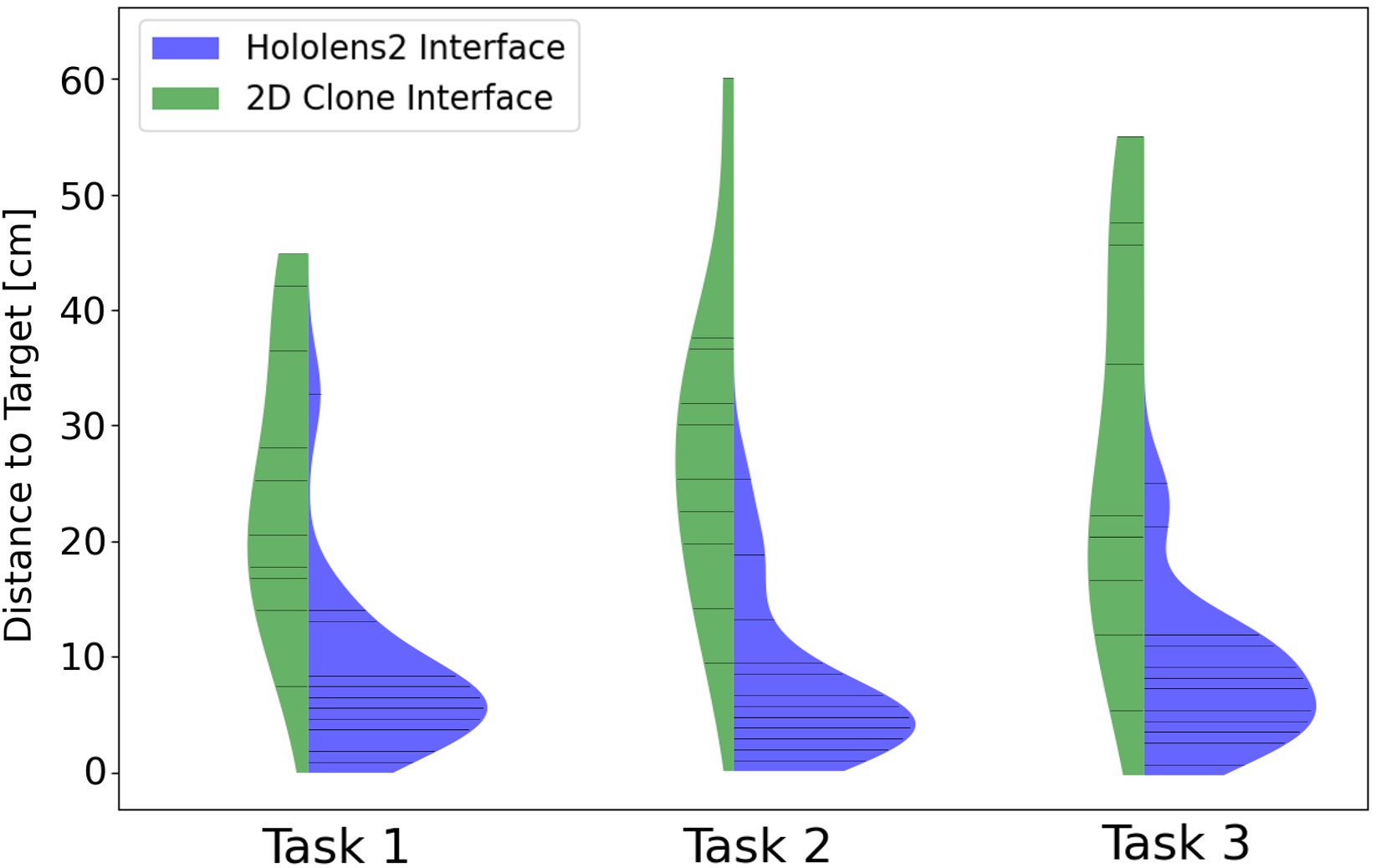}
    \vspace{-1mm}
    \caption{\textbf{User Accuracy Distribution Comparison. }Comparison of the participants' accuracy at placing the objects for the three tasks. Users with our proposed Hololens interface (blue) have significantly lower error while placing objects through drag-and-drop than users who use the 2D clone with mouse and keyboard (green). Horizontal lines mark the values for specific participants.}
    \label{fig:acc-comparison}
    \vspace{-0.5cm}
\end{figure}

\textbf{Success rate per object:} We use 6 kind of objects for testing the different difficulties. As shown in Table~\ref{tab:manipulation_stats_table}, the green watering can was the most reliably grabbed object and the cow plush animal has the lowest success rate of 55\%.
Different objects caused different failure modes. The green watering can, for example, collided on navigation and destination since the robot doesn't take the carried object into account for local collision avoidance. The mug is specially small, causing many empty grasps. Finally, for the stuffed cow, the main issue was due to collision during grasp. This is likely because the cow was placed in the most cluttered pick-up location. We conclude that there systematic failures for individual objects in relation to our proposed interface. It is likely that more advanced mobile manipulation systems could overcome most observed failure cases.

\subsection{User Study}

We designed a user study to assess the intuitiveness and usability of our interface. The study was performed on a varied group of 24 participants, aging 20 to 66 years old, with technical and non-technical backgrounds. None of them had used the Hololens before. The study has been approved by the ETH Ethics Commission under project number 268.  For this study, a simulated environment is used to maintain controlled experimental conditions. In the first part of the study, the participants are asked to complete the app's tutorial and are given time to familiarize themselves with the controls. In the second part, the users have to perform a set of three tasks. These tasks consisted of moving specific objects to different locations marked by visual indicators (e.g., 'Place the object on top of this shelf where you see the cross'). The third task is to command the robot to open a drawer and place a cup inside the opened drawer. Task completion rate and user accuracy are recorded for each task. User accuracy is measured as distance between predetermined goal coordinates and the actual coordinates sent by users to the robot. Furthermore, the time for completing the tutorial and finishing the tests is recorded. Finally, after the tutorial and after completing the tests, the participants are asked to respond to questions from both the NASA Task Load Index (TLX)~\cite{hart2006nasa} and the System Usability Scale (SUS) questionnaire~\cite{SUS-questionnaire}, including questions on frustration, mental demand and ease-of-use. 

To compare the performance of the proposed Hololens setup to a traditional 2D interface for mouse and keyboard, a subsection of these users (10) also tested a 2D clone of the Hololens app using the Unity app simulator. These participants were asked to perform the same three tasks of the Hololens study on a monitor using mouse and keyboard. Duration, success rate, user accuracy, and the NASA TLX questionnaire answers were also recorded on this test.




\begin{table}[b]
\vspace{-3mm}
\centering
\resizebox{1\columnwidth}{!}{
\begin{tabular}{clcc|cc}
\toprule
& & \multicolumn{2}{c|}{\textbf{Hololens}} & \multicolumn{2}{c}{\textbf{Mouse \& Keyboard}} \\
& \textbf{Measure} & \textbf{Mean} & \textbf{Std Dev} & \textbf{Mean} & \textbf{Std Dev} \\
\midrule

\multirow{3}{*}{\raisebox{-.5\height}{\rotatebox{90}{Tutorial}}} 
& Duration & 03:07 & 1:35 & - & - \\
& Perceived Frustration [1-10]& 4.43 & 1.91 & - & - \\
& Perceived Effort [1-10]& 3.71 & 1.52 & - & - \\
\midrule
\multirow{6}{*}{\rotatebox{90}{Tasks 1-3}} 
& Duration & 05:56 & 02:02 & 04:47 & 1:42 \\
& Test 1 Accuracy [cm] & 8.29 & 7.27 & 27.1 & 13.3 \\
& Test 2 Accuracy [cm] & 8.57 & 6.97 & 31.6 & 15.29 \\
& Test 3 Accuracy [cm] & 10.14 & 6.54 & 31.2 & 17.89 \\
& Perceived Frustration [1-10]& 3.76 & 2.43 & 6.5 & 1.77 \\
& Perceived Effort [1-10]& 4.43 & 1.83 & 6.7 & 1.42 \\
\midrule
\multirow{2}{*}{\rotatebox{90}{All}} 
& Subjective Performance [1-10]& 6.90 & 1.84 & - & - \\
& Ease-of-Use [1-5]& 3.57 & 0.93 & - & - \\
\bottomrule
\end{tabular}
}
\caption{\textbf{User Study Results.} Tutorial, tests, and overall measures. Comparison between the proposed Hololens interface and the 2D clone that uses mouse and keyboard.}
\label{tab:study_results}
\vspace{-0.4cm}
\end{table}

The main results are shown in Table~\ref{tab:study_results}. The average participant finished the tutorial in 2 to 4 minutes. It served as a short introduction to our proposed interface and to the Hololens in general. New users do not feel overwhelmed while completing the tutorial, on average the perceived effort to finish the introductory tasks is a 3.7. Furthermore, as the participants get more familiar with the controls, their frustration levels decrease from 4.4 after the tutorial to 3.8 at the end of the study. Participants in the end felt comfortable with the interface, rating their own performance in the upper range of the scale (6.9). These results all indicate the high usability of our app, even for users who are new to the HoloLens or to virtual reality.

When comparing these results with the participants who also tested the 2D app clone, the object misplacement of the 2D counterpart is more than 3 times as high. Figure~\ref{fig:acc-comparison} visualizes these results in more detail. The majority of participants using the Hololens are able to maintain the distance to the target under 10cm, while the 2D app clone results span from 10cm to half a meter. The main cause for the error is the user not being able to correctly estimate the depth of the object's location in the monitor. This is also results in participants reporting being more frustrated with the 2D app clone interface. The interface comparison underlines the effectiveness of our proposed interface and shows that augmented reality platforms are better suited for placement of objects in 3 dimensional spaces.

\section{CONCLUSION AND FUTURE WORK}

We introduced a novel interface that enables intuitive pick and place commands for a Spot robot through a drag-and-drop in a 3D MR environment. Our solution offers an interactive and user-friendly app including a tutorial. The conducted user study shows that users without experience with MR devices can successfully operate the robot and are 3 times more accurate in object placing compared to traditional mouse and keyboard interfaces. 
Our experiments show that the investigated overall system allows operators to command pick-and-place tasks on a variety of objects and in different environments with a total success rate of 72\%.

Possible future improvements include: Refining the drag-and-drop mechanics based on the object type, enabling the system to perform more sophisticated tasks like turning on switches, and replacing the manual list of movable object categories with a better prior e.g. from a LLM.

{
\newpage
\bibliographystyle{ieee_fullname}
\bibliography{sections/references}
}

\end{document}